\documentclass[10pt,journal,compsoc]{IEEEtran}

\usepackage{graphicx}
\graphicspath{{figures/}}
\usepackage[nocompress]{cite}
\usepackage{amsthm}
\usepackage[T1]{fontenc}

\usepackage{tikz}
\usepackage{booktabs} 
\usepackage{xspace}
\usepackage{subfig}
\usepackage{array}
\usepackage{multirow}
\usepackage{amsmath}
\usepackage{amssymb}
\usepackage[ruled,vlined, linesnumbered]{algorithm2e}
\usepackage{setspace}

\newcommand{\paratitle}[1]{\vspace{1.5ex}\noindent \textbf{#1}}
\newcommand{\ie}{\emph{i.e.,}\xspace}
\newcommand{\eg}{\emph{e.g.,}\xspace}
\newcommand{\etal}{\emph{et al.}\xspace}

\newtheorem{mydef}{Definition}

\begin{document}
\title{UFTR: A Unified Framework for Ticket Routing}

\author{Jianglei Han,
	Jing Li
	and Aixin Sun
	\IEEEcompsocitemizethanks{\IEEEcompsocthanksitem J. Han is a part-time Ph.D student at School of Computer Science and Engineering, Nanyang Technological University, Singapore. E-mail: jhan011@e.ntu.edu.sg
		
		\IEEEcompsocthanksitem J. Li is with Inception Institute of Artificial Intelligence, United Arab Emirates. E-mail: jingli.phd@hotmail.com. The work was done when the author was with Nanyang Technological University, Singapore.
		
		\IEEEcompsocthanksitem A. Sun is with School of Computer Science and Engineering, Nanyang Technological University, Singapore. E-mail: axsun@ntu.edu.sg.}
}


\IEEEtitleabstractindextext{%
\begin{abstract}
Corporations today face increasing demands for the timely and effective delivery of customer service. This creates the need for a robust and accurate automated solution to what is formally known as the \textit{ticket routing problem}. This task is to match each unresolved service incident, or  “ticket”, to the right group of service experts. Existing studies divide the task into two independent subproblems – initial group assignment and inter-group transfer. However, our study addresses both subproblems jointly using an end-to-end modeling approach. We first performed a preliminary analysis of half a million archived tickets to uncover relevant features. Then, we devised the \textbf{UFTR}, a \textbf{U}nified \textbf{F}ramework for \textbf{T}icket \textbf{R}outing using four types of features (derived from tickets, groups, and their interactions). In our experiments, we implemented two ranking models with the UFTR. Our models outperform baselines on three routing metrics. Furthermore, a post-hoc analysis reveals that this superior performance can largely be attributed to the features that capture the associations between ticket assignment and group assignment. In short, our results demonstrate that the UFTR is a superior solution to the ticket routing problem because it takes into account previously unexploited interrelationships between the group assignment and group transfer problems.
\end{abstract}
}

\maketitle
\IEEEdisplaynontitleabstractindextext
\IEEEpeerreviewmaketitle

\section{Introduction}
\IEEEPARstart{M}ore companies are embracing Information Technology (IT) in their operations~\cite{galup2009overview}. To standardize IT-related activities, the British government developed the IT Infrastructural Library (ITIL), which defines the standards and best practice across the entire IT service life-cycle (\eg strategy, design, transition, operation, and continuous service improvement). According to the ITIL, “operation” refers to processes which are crucial to the day-to-day running of companies, encompasses the direct delivery of goods and services from service providers to end users. Our study investigates the \textit{ticket routing problem} in incident management and service operation.

\begin{figure}
	\centering
	\includegraphics[width=0.85\linewidth]{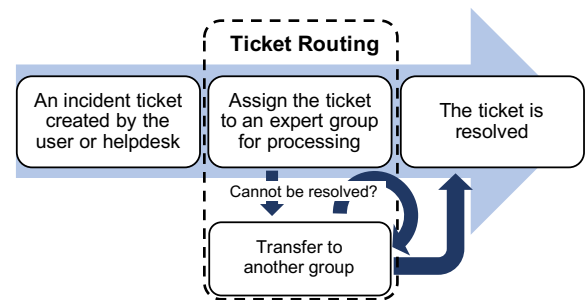}
	\caption{Incident management workflow as recommended by ITIL. The Ticket Routing problem pertains to the middle of the workflow (enclosed by dotted lines), including assigning the ticket to an initial expert group for processing, and transferring it to another group as if it cannot be resolved.}
	\label{fig:routing}
\vspace{-0.5cm}
\end{figure}

An incident is an event that prevents a user from performing their task. It could be due to a system fault, an access request, or a lack of user knowledge. An \textit{incident ticket} is a document containing user-generated text and system information. Figure~\ref{fig:routing} depicts a standard workflow for processing a ticket, recommended by ITIL. First, the user creates an incident ticket, either directly or by contacting the helpdesk. Each newly created ticket needs to be matched with an expert group in charge of processing it. An \textit{expert group} is a unit of supporting staff who are experts in certain areas. If the first assigned group solves the problem, the ticket is resolved (or closed). Otherwise, the routing system needs to transfer the ticket to another group for processing until the ticket is resolved. Initial group assignment and inter-group transfer are individually studied as problems of \textit{routing recommendation}~\cite{Shao2008EasyTicket:Resolution,Miao2010GenerativeNetworks,Moharreri2016MotivatingManagement}, and \textit{expert recommendation}~\cite{Xu2018ExpertRouting,Motahari-Nezhad2011}. In this work, we consider both collectively as parts of the larger \textit{ticket routing problem}.

Manual ticket routing relies heavily on the human experience and predefined rules. Incorrect routing results in delay of incident resolution and the waste of processing resources. Previous works~\cite{BShivaliAgarwalRenukaSindhgatta2012,Zeng2017KnowledgeData,Han2018VerticalNetworks} approach the problem using classification models trained only on textual content of the tickets. Groups that successfully resolve the closed tickets are positive samples for training the classifiers. Using ticket transfer histories between networks of expert groups, Shao~\etal~\cite{Shao2008EfficientMining,Shao2008EasyTicket:Resolution} proposed Markov models for routing without concerning ticket contents. Subsequent works~\cite{Khan2009AIM-HI:Delivery,Miao2010GenerativeNetworks,Sun2010} propose models using both ticket contents and transfer histories. Recently, Xu and He~\cite{Xu2018ExpertRouting} propose a framework for expert assignment and ticket transfer in two stages, in which different models are trained independently. 

In our study, we propose a joint framework that takes features from both expert assignment and ticket transfer in an end-to-end model.
We study a large support organization consisting thousands of expert groups, which process hundreds of thousands of tickets annually. Through a large-scale analysis on archived tickets, we find that tickets with shorter routing sequences have higher content clarity (or topic specificity). We also find that the probability of playing specific roles in ticket routing, \eg resolver or non-resolver, differs across different types of expert groups. As resolvers, the source groups from which they receive the tickets from are different are modeled with different probability distributions. We extract network features by modeling the expert groups using different structure information (\eg group hierarchy). Such structures are commonly seen in large support organizations, but were not considered for similar problems previously.

Based on our preceding analysis, we propose the UFTR, a \textbf{U}nified \textbf{F}ramework for IT \textbf{T}icket \textbf{R}outing. For each ticket-group pair, it computes a matching score. The features models the individual and interactive information from the ticket and expert groups. It is unified and extensible, incorporating findings from most of the previous studies as features. We categorize the features into four types: features derived from tickets (T) and groups (G) respectively, \ie ticket features and group features, and features derived their interactions, \ie Ticket-Group (TG) features and Group-Group (GG) features. We implement UFTR with both pointwise and listwise ranking models, using the four types of features. Results show UFTR outperforms all baselines on processing real ticket data, evaluated by three routing evaluation metrics. Between the two ranking models, results produced by listwise model is marginally better. We found that TG is the most important to our ranking model among the four types of features. This finding differs from that of ~\cite{Xu2018ExpertRouting}, in which the authors concluded that ticket features are most important.

The main contributions of this paper are as follows. 
\begin{itemize}
	\item We analyze half million tickets from a large support organization. We share insightful findings we believe could be interested by researchers and practitioners working on a similar problem. 
	\item We formulate the ticket routing problem as a learning-to-rank problem. From the data, we generate four types of features for ranker training. Each type represents a distinctive aspect of information. Some of the features can be pre-computed offline (\eg G, GG), while the others need to be computed at run-time. All feature types are extensible with additional information. 
	\item We propose the UFTR, a novel ticket routing framework unifying the two sub problems, \ie initial group assignment and inter-group transfer. To the best of our knowledge, UFTR is the first unified model for the ticket routing problem.
	\item Through experiments, we show that the UFTR outperforms baselines, while the UFTR with pairwise ranker is more effective compared to its pointwise counterparts. Our experimental results also support that, as far as the ranking models are concerned, the most impactful features are from the Ticket-Group type.
\end{itemize}

The remaining of this paper is organized as follows: Section~\ref{related works} summarizes previous works related to our research. Section~\ref{preliminaries} introduces routing in hierarchical organizations and our problem statement. Section~\ref{data analysis} presents comprehensive data analysis, with respect to the  proposed feature types. In Sections~\ref{framework} and~\ref{experiement}, we propose the UFTR framework and compare it with baseline models. We conclude this paper in Section~\ref{conclusion} with discussion on future works.
              
\section{Related Work} \label{related works}

Earlier efforts at solving the ticket routing problem utilized graphical, generative, classification, and retrieval methods. We first summarize existing routing approaches, before discussing related work in ticket classification and retrieval.

\begin{table}[t]
	\small
	\centering
	\caption{Comparison of UFTR with existing routing systems in terms of feature types and models. The columns \textbf{Assign} and \textbf{Trans}, corresponds to ``initial group assignement" and ``inter-group transfer", respectively.}
		\scalebox{0.96}{
	\begin{tabular}{l |l l l l| c c}
		\toprule
		\textbf{System} & \textbf{T} & \textbf{G} & \textbf{TG} & \textbf{GG} & \textbf{Assign}  & \textbf{Trans}\\
		\midrule
		EasyTicket~\cite{Shao2008EasyTicket:Resolution} &-&-&-& $\checkmark$ & - & $\checkmark$ \\
		AIM-HI~\cite{Khan2009AIM-HI:Delivery} & $\checkmark$ &-&-&- & $\checkmark$  & -\\
		Hybrid model~\cite{Sun2010} & $\checkmark$ &-&-& $\checkmark$ &  $\checkmark$  & -\\
		GenerativeModel~\cite{Miao2010GenerativeNetworks} & $\checkmark$ &-& $\checkmark$ & $\checkmark$ & - & $\checkmark$\\
		ITSCM~\cite{Motahari-Nezhad2011} & $\checkmark$ &-&-& $\checkmark$ & $\checkmark$  & - \\
		TER~\cite{Xu2018ExpertRouting} & $\checkmark$ & $\checkmark$ & $\checkmark$ & $\checkmark$ & \multicolumn{2}{c}{Two stages}\\
		\midrule
		\textbf{UFTR (this work)} & $\checkmark$ & $\checkmark$ & $\checkmark$ & $\checkmark$ & \multicolumn{2}{c}{Unified}\\
		\bottomrule
	\end{tabular}}
	\label{tab:summary}
\end{table}

\paratitle{Ticket routing}. One of the first formulations of the problem was from Shao \etal~\cite{Shao2008EfficientMining,Shao2008EasyTicket:Resolution}. The main idea was to apply frequent pattern mining techniques to find transfers patterns in archived routing sequences. The extracted patterns are used for more efficient routing in Markov models. In such models, each expert group is modeled as a graph node, where the edges are normalized frequencies of ticket transfers between two nodes. The frequent patterns are used to select the next routing destination. Specifically, they proposed the most probable routing destination based on a few previous steps. For example, the model could choose a path \texttt{B} $\rightarrow$ \texttt{D} over \texttt{B} $\rightarrow$ \texttt{C}, even though the latter had higher relative frequency. The reason is, \texttt{[A, B]} $\rightarrow$ \texttt{D} has the highest relative frequency among all possible destinations, where \texttt{A} is a group before \texttt{B} in the routing path. The authors also investigated the usefulness of intermediate transfer groups in the routing process. They concluded that some groups act as distributors and they are important to final routing to resolver. However, the information-rich ticket contents were ignored.

Subsequent works attempted to integrate ticket contents in routing models. Sun \etal~\cite{Sun2010} proposed a multi-stage content-aware model. It first filters the expert groups based on content similarities to the input ticket, then builds a graph with a smaller number of nodes. The performance of such model was superior to the sequence-only models. Miao \etal~\cite{Miao2010GenerativeNetworks} proposed a generative Optimized Network Model (ONM), leveraging information from both ticket contents and historical routing sequences. It captures the hidden connections between terms from tickets and the transition probabilities to the next group. Compared to the multi-stage model, ONM was able to consider all other groups as candidates, but has a large parameter space for optimization. Both models require an initial group as input together with the ticket.  

More investigations~\cite{Chen2010AssessingNetworks,Miao2012UnderstandingNetworks,Sun2014AnalyzingNetworks} into the ticket transfer patterns were made, in slightly different settings and datasets. It was found that expert groups' expertise and awareness of others' profiles are the main factors contributing to decision making. While the former information is relatively straightforward to obtain, the latter is difficult to model in real organizations.

Recently, Xu and He~\cite{Xu2018ExpertRouting} proposed a Two-stage Expert Routing (TER) model for ticket routing problem. In the first stage, the initial group is determined by taking the nearest neighbor in a distributed expert group representation space, as per~\cite{Han2016DistributedExpertise}. Assuming a group only handles tickets with matching skills, low dimensional vectors representing groups are learned using tickets from archived collections. Once trained, a new ticket can be projected to the same vector space as the expert groups for distance-based selection. In the subsequent routing stage, the authors employed probabilistic transition models from~\cite{Shao2008EfficientMining}, ignoring ticket contents during the process. In short, for each stage, the TER framework trained a model independently from each other. As a result, the rich interaction information contained in ticket routing histories are under-utilized. With this in mind, we propose the UFTR to handle both stages with a unified model, leveraging full sets of accessible information.

\paratitle{Ticket classification}. Without considering inter-group transfers, ticket routing can be reduced to a text classification task, in which the resolver group is the target label. Previous works explored a large number of different features to use with standard classifiers, \eg Support Vector Machines (SVM), Na\"ive Bayes (NB), Random Forest etc. Khan \etal~\cite{Khan2009AIM-HI:Delivery} extracted features from a set of pre-defined signatures from tickets before applying various supervised algorithms to predict the resolver. Agarwal \etal~\cite{BShivaliAgarwalRenukaSindhgatta2012} proposed SmartDispatch, using a combination of Support Vector Machines (SVM) and discriminative term-based classifications for resolver determination. The system design considered the cases where predictive models alone are not able to handle. Specifically, when a ticket has the prediction confidence lower than a threshold score, SmartDispatch transfers it to a human dispatcher for processing. This solution focused on initial group assignment, instead of routing. In another series of works, Moharreri \etal~\cite{Moharreri2015,Moharreri2016ProbabilisticNetworks,Moharreri2016MotivatingManagement} investigated several multi-level classification systems, before recommending a routine sequence that had previously met the service level time requirement. As our objective is to identify the resolver, recommending a sequence of groups is not applicable accroding to our problem setting. There were also works categorizing tickets into domain-specific categories~\cite{Diao2009,Zeng2017KnowledgeData,Han2018VerticalNetworks}. In general, these applications differed only in features (\ie to represent tickets) and labels (\ie to represent classification targets). Off-the-shelf classification algorithms were usually applied. 

\paratitle{Ticket retrieval}. Ticket retrieval is related to ticket routing, since it is about ticket representation and similarity measures among tickets. Specifically, in case-based reasoning (CBR) framework~\cite{Simoudis1992,Kang2014AKnowledge}, resolved tickets are stored and managed in a ``case-base'', together with their associated resolvers, or solutions. For a new ticket, the task is to find its most likely resolver, or solution, by searching the case-base. In the process, the new ticket is a query. With  similar tickets returned from the case-base, the resolvers or resolutions are aggregated and ranked. Traditionally, standard text similarities were used for similar tickets retrieval, \eg BM25, term frequency-inverse document frequency (TF-IDF), and Jaccard similarity. Netzhad \etal~\cite{MotahariNezhad2011AnalyticsFlows} considered the matching of activity sequences and expert profiles in addition to text features. To improve the effectiveness of similarity search, Kang \etal~\cite{Kang2014AKnowledge}  incorporated external knowledge in addition to ticket contents. Specifically, the authors modeled incident-classification taxonomy, workgroup taxonomy, and keyword-class taxonomy for knowledge-rich pairwise ticket similarity matching. These similarities can be extended to measure the similarity between a ticket and a group, where the group is represented by the tickets it has resolved.  

Table~\ref{tab:summary} summarizes existing routing systems in terms of the features and tasks. One can observe that most of these models leverage partial information from the feature types we identified, except TER. In comparison, UFTR uses all types of features to better capture the individual and interaction information between tickets and expert groups. Note that UFTR is different from TER, as the latter uses two-stage models, where UFTR is a unified framework.
UFTR incorporates findings from previous studies in ticket routing, classification, and retrieval. Ticket features are researched in all these tasks, which are then generalizabled to a few representative features. In addition, we explore network features for expert groups as well, and further Ticket-Group and Group-Group interaction features.

\section{Preliminaries} \label{preliminaries}

\begin{figure}[t]
	\centering
	\includegraphics[width=\linewidth]{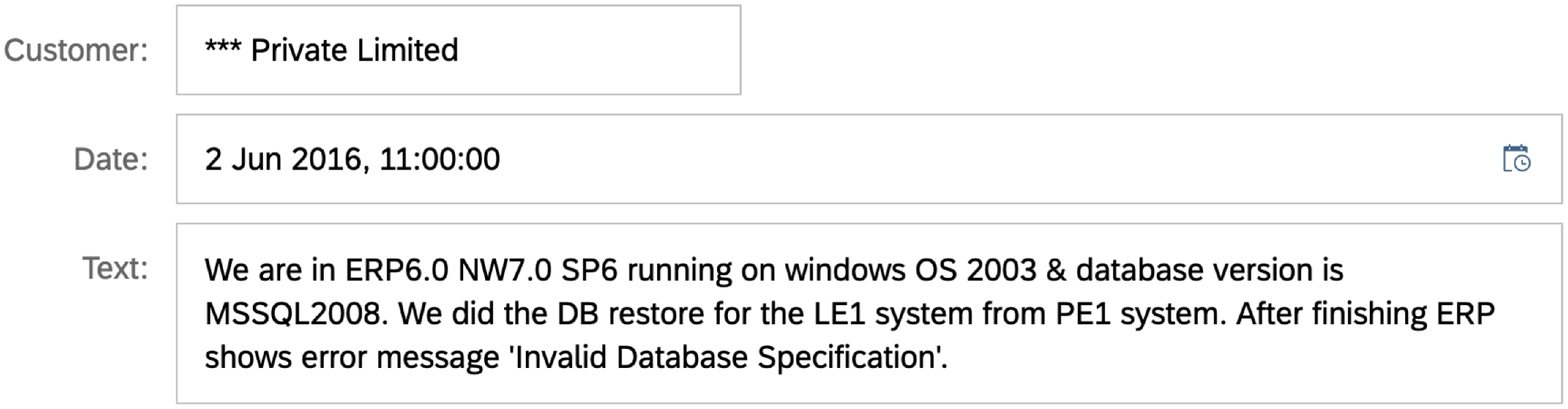}
	\caption{An illustration of an incident ticket. The customer information is anonymized.}
	\label{fig:sample}
	\vspace{-0.3cm}
\end{figure}

In Fig~\ref{fig:sample}, we present a sample ticket for illustration purpose. As shown, it consists of structured information (\eg customer name, date and time the ticket was created), and unstructured text description of the incident. Due to the requirements of a data protection agreement, we only extract the text field from each ticket for analysis and experimentation. In addition, time information are not within the scope of this research.

\begin{mydef}[Ticket]
	A ticket is a text document consists of incident description.
\end{mydef}

\begin{figure}[t]
	\centering
	\includegraphics[width=0.8\linewidth]{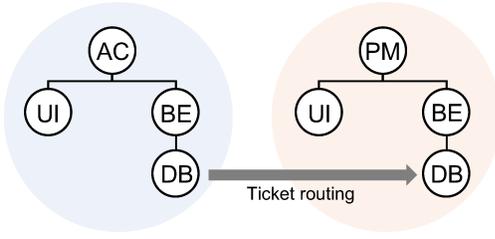}
	\caption{Illustration of a ticket routing process. \texttt{AC} is the root for ``Accounting" products, while \texttt{PM} is the root for``Process Management" products. A sample ticket is first assigned to  \texttt{AC$\cdot$BE$\cdot$DB}, which is unable to resolve the issue. The ticket is routed to \texttt{PM$\cdot$BE$\cdot$DB}, which turns out to be the resolver group. Here, \texttt{BE} and \texttt{DB} are ``back-end" and ``database" components of their root products, respectively.}
	\label{fig:hierarchy}
	\vspace{-0.3cm}
\end{figure}

In the organization we studied, expert groups are the basic units of ticket processing. We consider a pre-defined collection of expert groups $\mathcal{G}$. They are organized by their area of responsibilities and expertise. Each functional area is represented in a tree of expert groups. We consider two levels of abstractions: at the $root$ level ($g^{root}$), where a root node is the collection of all its children groups; and the $leaf$ level ($g^{leaf}$) which is the group itself. In the remaining of this paper, we use \texttt{AA$\cdot$BB$\cdot$CC} to denote an expert group, where $\cdot$ represents a child-of relationship, and \texttt{AA} is a placeholder for root node. For example, in Figure~\ref{fig:hierarchy}, two group trees have \texttt{AC} and \texttt{PM} as roots, respectively. A routing is taking place at leaf level from group \texttt{AC$\cdot$BE$\cdot$DB} to  \texttt{PM$\cdot$BE$\cdot$DB}. At root level, the routing is from  \texttt{AC} to \texttt{PM}.

\begin{mydef}[Expert group]
	An expert group is the ticket processing unit. Each root is independent of other, representing a major functional area. Within a root, child nodes are hierarchically structured. Each group is responsible for a subset of incident tickets. The group that closes a ticket is the resolver group.
\end{mydef}

A routing sequence is a list of expert groups that are involved in the routing process of a ticket. The group in each step of the routing is a participant group ($g_{participant}$). A participant group may play different roles when it is involved in processing a ticket. The first group in a routing sequence is the initial group ($g_{initial}$) a ticket is assigned to. The group that closes the ticket is the resolver ($g_{resolver}$). A ticket may take only one routing step to close; that is, the initial and resolver groups are the same group, and the group is also a participant by definition.

\begin{mydef}[Routing sequence]
	A routing sequence consists of one or more expert groups which handled the tickets, starting with the initially assigned group. The last group in the sequence is the resolver group.
\end{mydef}

\paratitle{Problem Formulation.} Instead of modeling initial group assignment and ticket routing as separate tasks, we consider \textit{ticket routing as a unified problem}. Specifically, given a ticket $\tau$ and its current routing sequence $S_{t}$ (if available), the task is to recommend a ranked list of expert groups for human to decide which next group to route to. $S_{t}$ contains all expert groups who have processed $\tau$, but failed to close it. When a new ticket is just created, its current routing sequence is empty, \ie $S_{t} = \emptyset$.

\section{Exploratory Analysis} \label{data analysis}
\begin{figure*}
	\includegraphics[width=\linewidth]{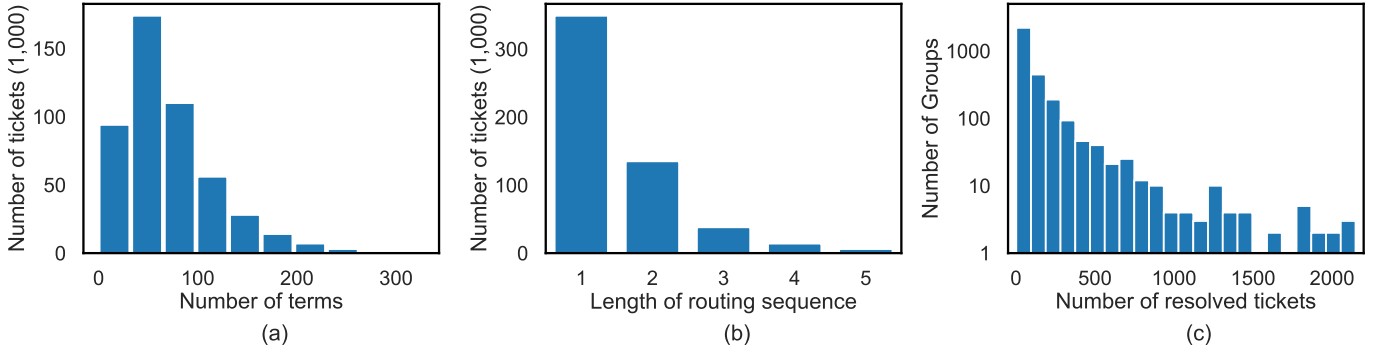}
	\caption{Overall statistics from 500,000 tickets. (a) ticket size distribution. (2) routing sequence length distribution. (3) Group density distribution by number of resolved tickets.}
	\label{fig:counts}
\end{figure*}

We randomly sample half a million archived tickets with their routing sequences from a support system. Figure~\ref{fig:counts} shows some basic data statistics. Figure~\ref{fig:counts}(a) shows ticket size distribution, by number of terms. The histogram shows a peak at around 50 terms, and the majority of tickets are shorter than 100 terms. Figure~\ref{fig:counts}(b) shows that the routing sequence length follows a Power Law distribution. More than 50\% of the tickets are resolved by the first assigned group without routing. Previous works~\cite{Shao2008EfficientMining,Miao2010GenerativeNetworks,Xu2018ExpertRouting} taking current processing group as input, do not support this majority of cases in our data. Figure~\ref{fig:counts}(c) shows the distribution of group density. The x-axis represents number of resolved tickets. The peak at left end shows most of the groups have resolved fewer than 1000 tickets. While at the right end, only a handful groups have resolved a lot of tickets.

Next,  we attempt to answer the following research questions through more detailed analysis.

\paratitle{RQ1}: Does clarity of a ticket correlate with the number of steps needed to reach the resolver?

\paratitle{RQ2}: Are some groups more likely to be resolvers?

\paratitle{RQ3}: For resolver groups, how often do they receive tickets from other groups, and/or directly from dispatch?

\subsection{Ticket Content Analysis}
A ticket contains rich information for human experts to diagnose the incident. In our data, we observe high usage of technical terms/concepts and variant representations of the same concepts and entities.

\paratitle{Technical Entities}. In an earlier study, Han~\etal~\cite{Han2018TowardsTickets} investigates software product name recognition and normalization in tickets. In this study, we extend the findings to a more general notion of \textit{technical entities}, including product names, technical terms and concepts. In sample ticket from Figure~\ref{fig:sample}, ``We are in \underline{ERP6.0 NW7.0 SP6} running on \underline{windows OS 2003} \& \underline{database version} is \underline{MSSQL2008}. We did the \underline{DB} restore for the \underline{LE1} system from \underline{PE1} system. After finishing \underline{ERP} shows error message...", the technical entities are underlined.

After determining the entities, tickets can be represented using an ``entity-oriented" model. Such representation has two benefits. First, technical entities are more semantically representative and the number of dimensions are reduced by an order of magnitude (\eg from 400 thousand to 40 thousand in our case). 
Second, sensitive personal information is automatically filtered out from the representation. This is a crucial requirement by General Data Protection Regulations (GDPR)~\cite{gdpr-web}. In this work, we assume the technical entities are extracted by upstream process, while the details of extraction are out-of-scope of this paper.

\paratitle{Technical Clarity (for RQ1)}. With  ``bag-of-technical entities" representation, we further develop a technical clarity score to evaluate topic specificity of a ticket. Cronen-Townsend~\etal~\cite{Cronen-Townsend2002PredictingPerformance} proposed a clarity score to predict query difficulty using the relative entropy between a query language model and the collection language model. We replace words with the technical entities in the original model to compute a technical clarity for a ticket $\tau$. Formally,
$$\text{Clarity}(\tau) = \sum_{e \in E} P(e|\tau)\log_2\frac{P(e|\tau)}{P(e|C)},$$
where $E$ is the set of entities in the collection and $C$ is the collection of tickets.

\begin{figure}
	\centering
	\includegraphics[width=0.85\linewidth]{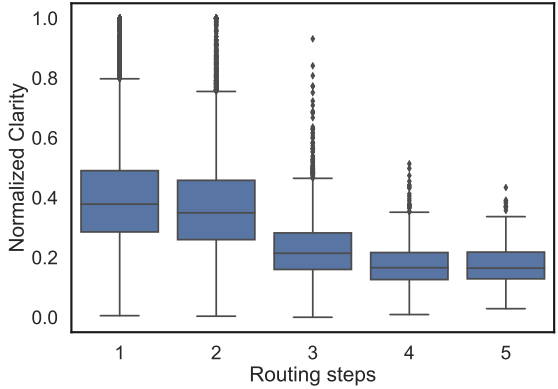}
	\caption{Distributions of clarity scores of archived tickets, grouped by the number of routing steps.}
	\label{fig:clarity}
\end{figure}

Figure~\ref{fig:clarity} plots the  distribution of technical clarity of tickets, grouped by their number of routing steps. Observe that tickets with shorter routing sequences have higher mean clarity than those with longer sequences, although the  variance is relatively higher.

\subsection{Expert Group Analysis}
\label{group and network analysis}

\begin{figure*}[t]
	\centering
	\includegraphics[width=0.85\linewidth]{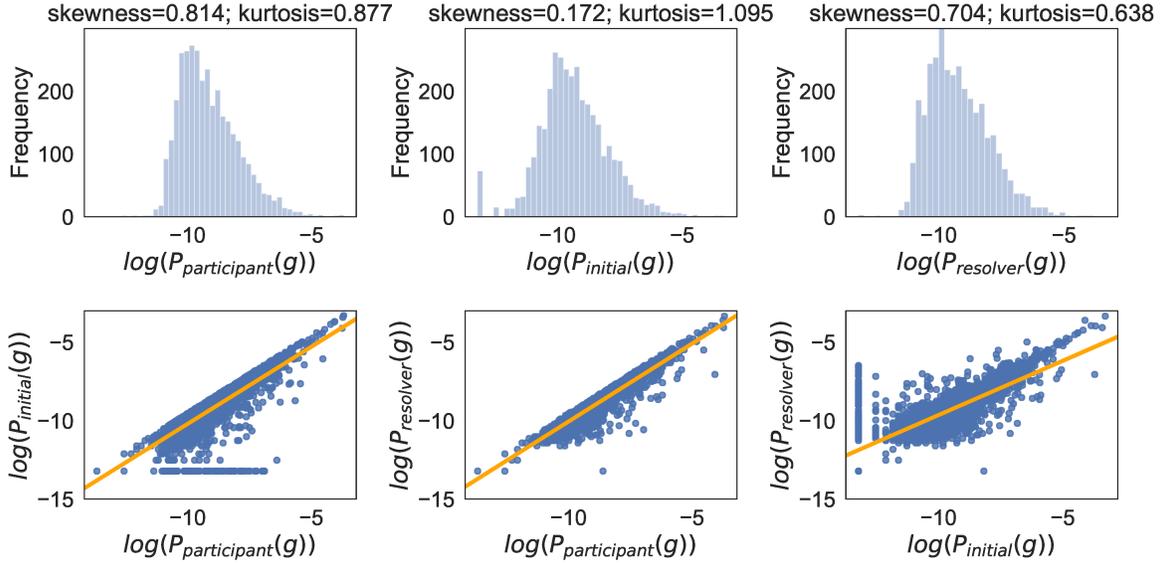}
	\caption{Log probability distribution of expert group being different roles, and the correlations of probability distribution of the same group playing different roles.}
	\label{fig:prob_resolver}
\end{figure*}

\paratitle{Statistical Analysis (for RQ2).} With a large number of expert groups in the network and transfer histories, we first characterize groups from a statistical perspective. Specifically, we investigate the probability distribution of different roles in transfer histories and study the pairwise correlations. We use three random variables to model the probability of a group as the initial group ($P_{initial}(g)$), the resolver ($P_{resolver}(g)$), or participant ($P_{participant}(g)$) in routing\footnote{Note that participant group is a group involved in a routing sequence. Both initial and resolver groups are also participant groups.}. The values are estimated by $$P_{initial}(g) = \frac{\#(g~\text{being the initial group})+1}{|\mathcal{T}| + |\mathcal{G}|},$$ $$P_{resolver}(g) = \frac{\#(g~\text{being the resolver group})+1}{|\mathcal{T}| + |\mathcal{G}|},$$ $$P_{participant}(g) = \frac{\#(\text{All tickets involves}~g)+1}{|\mathcal{T}| + |\mathcal{G}|}$$ where $|\mathcal{T}|$ and $|\mathcal{G}|$ are total number of tickets and groups, respectively. A smoothing factor $1$ is added to prevent the result to be $0$. In the first row of Figure~\ref{fig:prob_resolver}, the log probabilities of these values are presented. The histograms show slight right-skewed normal distributions for all three variables. For each variable, we compute skewness and kurtosis, which are statistical descriptors for measuring the symmetry and tailedness of a distribution with respect to a normal distribution. Among the three variables, $log(P_{initial}(g))$ is the least skewed but has highest kurtosis, due to infrequent deviations or outliers in the data distribution.

In the second row, it shows pairwise correlations of probability distributions of the same expert group acting in different roles. Both $x$- and $y$-axis corresponds to expert groups sorted by their probability of being involved in a routing process.
We observe a stronger correlation of \textit{initial-participant} and \textit{resolver-participant}, compared to \textit{resolver-initial}.
In \textit{initial-participant} and \textit{resolver-participant}, the data points below the regression line corresponds to  groups which have lower probabilities of being the initial group or resolver, than being involved in routing. Remarkably, in \textit{initial-participant}, there is a cluster of groups that have much lower probabilities to be the initial group than they would be involved in any role. The same pattern does not prevail in \textit{resolver-participant} subplot. Quantitatively, this difference is reflected by the higher kurtosis of $log(P_{initial}(g))$ compared to $log(P_{resolver}(g))$.
In \textit{resolver-initial}, the data points are scattered over a wider area on both sides of the regression lines. This shows some groups are more likely to be the resolver than to be the initial group, and vice versa. Note that the spike at the left end of the $x$-axis, there are groups that are rarely the initial group, but more likely to be the resolver.

\begin{figure}[t]%
	\centering
	\subfloat[Distribution of leaf groups]{{\includegraphics[width=0.46\linewidth]{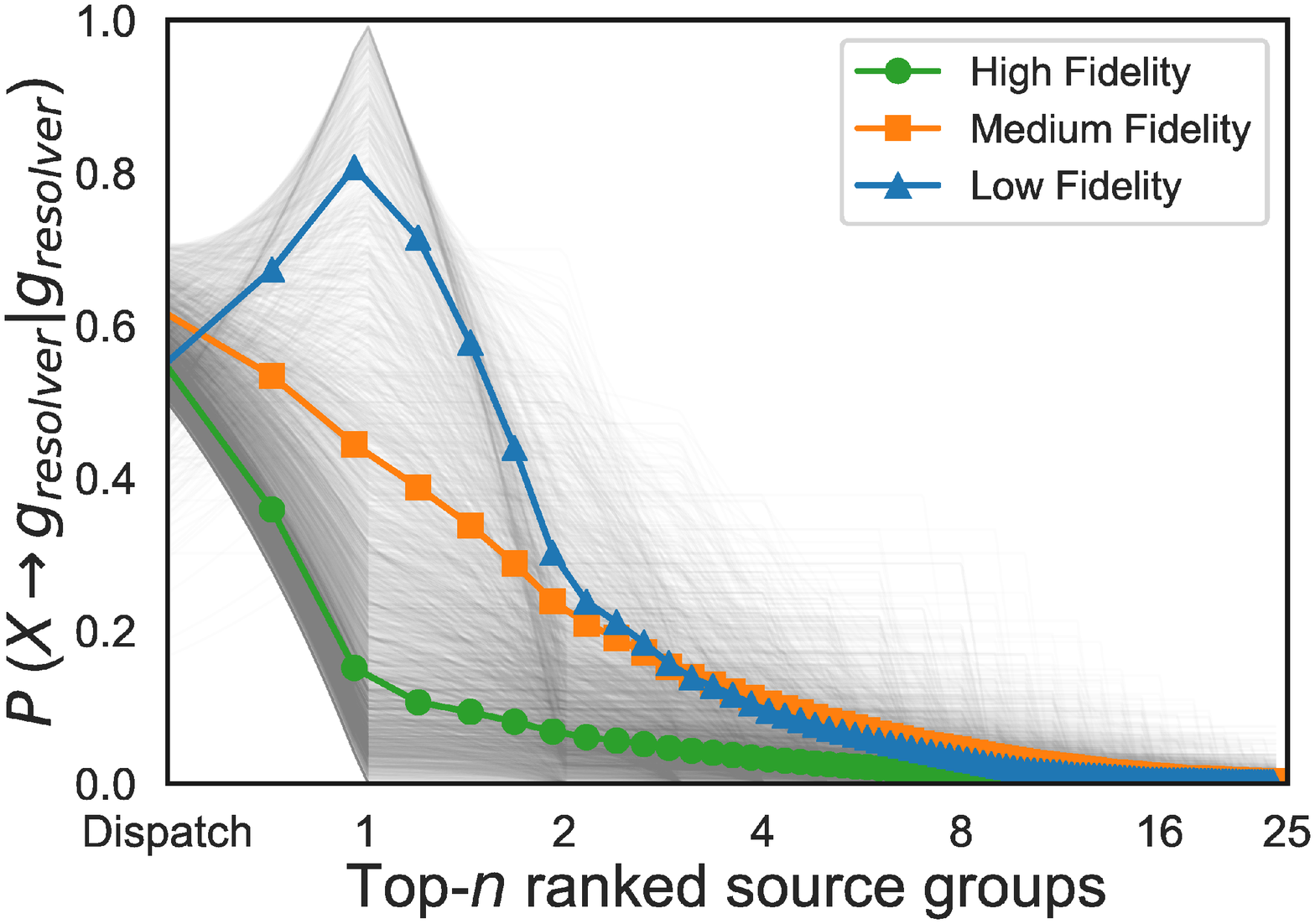}}\label{fig:source1}}%
	\qquad
	\subfloat[Distribution of root groups]{{\includegraphics[width=0.46\linewidth]{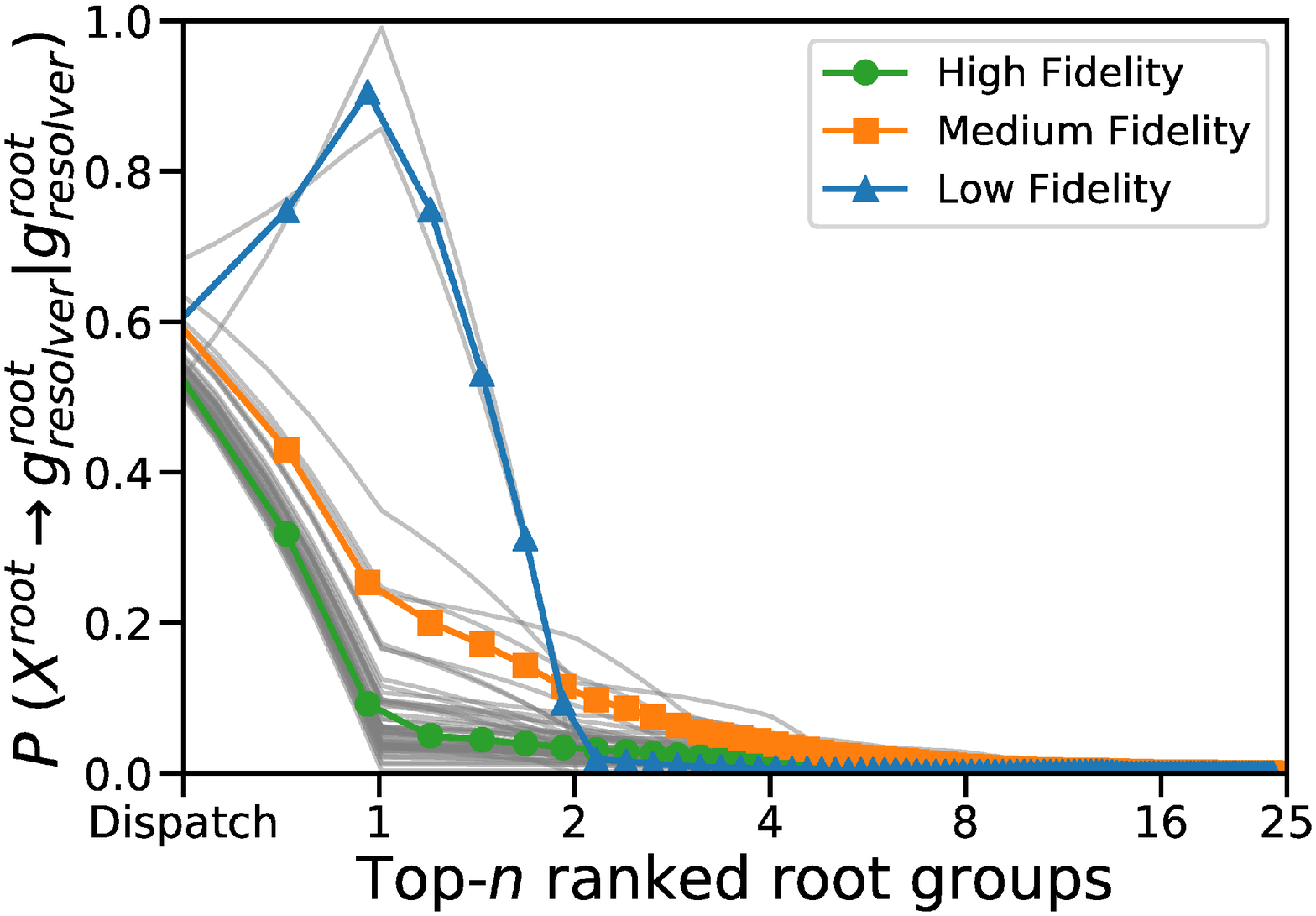}}\label{fig:source2}}%
	\caption{Source to resolver analysis. The left end of the $x$-axis is fixed for routings from dispatch. Each line characterize one expert group with the distribution of cumulative sources to resolver. The colored lines are the cluster centers (best viewed in color).}%
	\label{fig:source}%
\end{figure}

\paratitle{Source to Resolver Analysis (for RQ3).}  Since our main objective is to recommend resolvers, we are interested in the statistical characteristics for routing to resolvers. From historical routing sequences, we extract and model each resolver group ($g_{resolver}$) as a probability distribution of the groups in the previous routing step~\ie source to resolver. Since majority of the tickets are solved in one step, we fix the first source for each resolver to be from ``dispatch" and estimate the probability. Similarly, we estimate the probability for each source and rank in descending order \textit{for each resolver}. The resultant distributions are shown in Figure~\ref{fig:source}. Observe that with ``dispatch" source fixed at the left end of $x$-axis, the source to resolver distribution exhibits distinctive shapes, which we refer to as \textit{fidelity}. We further cluster the distributions using the $k$-means algorithm into 3 clusters, and show the cluster centers in colors. A large majority of the resolvers receive tickets from mainly one source, corresponding to a power law cluster center distribution with steep slop. We label this cluster as \textit{high fidelity} (green line). The second cluster has \textit{medium fidelity}, showing a smooth distribution of sources over the rank (orange line). Lastly, there is a cluster of resolvers with \textit{low fidelity} (blue line). That is, they receive tickets mainly from a source other than from dispatch.

To understand the distribution patterns at the root level (see Figure~\ref{fig:hierarchy}), we repeat the process for the root group ($g_{resolver}^{root}$). As shown in Figure~\ref{fig:source2}, only two root groups still show low fidelity. An important finding is that a majority of the resolvers receive tickets from dispatch or \textit{another group within the hierarchy with the same root}. Further investigation reveals close business connection between the two low fidelity groups (finance and accounting). This shows that groups from different roots are also connected.

\subsection{Network Analysis} 
\label{network analysis}

\begin{table}[t]
	\centering
	\caption{Statistics and characteristics of networks $H_{trans}$, $H_{res}$, and $H_{root}$.}
	\begin{tabular}{lrrr}
		\toprule
		& $H_{trans}$ & $H_{res}$ & $H_{root}$ \\
		\midrule
		\# of nodes & 3,390  & 3,390 & 57\\
		\# of edges  & 44,719 & 82,904 & 1,390\\
		Network Diameter & 10 & 8 & 3\\
		Average degree & 14.161 &  24.455 & 24.123\\
		Average weighted degree & 0.996 & 0.735 & 0.599\\
		Average path length & 3.299 & 2.978 & 1.592\\
		Average clustering coefficient & 0.218 & 0.268 & 0.722\\
		\bottomrule
	\end{tabular}
	\label{tab:network}
\end{table}

To understand the inter-group connections, we define three directed graph $H = (\mathcal{G}, R)$ to model the group interaction, where $R$ is the set of edges. The first graph, $H_{Trans}$, is modeled following~\cite{Shao2008EfficientMining}: an edge $g_i \rightarrow g_j$ means ticket transfer from $g_i$ to $g_j$ and the edge weight is the fraction of tickets from $g_i$ to $g_j$ with respect to all out-going tickets from $g_i$. However,   $H_{Trans}$ does not differentiate the role of resolver in routing, as edges correspond to all routing pairs.  We argue that all routing groups in a routing path are related to resolver to different extent. To this end, we  construct another network, $H_{Res}$ to model the distance between a group to resolver, in routing sequences. In particular, for a routing path, we connect all previous routing groups to the resolver. The edge weights decrease as further away the group from the resolver. Meanwhile, we also remove edges among previous groups (as our goal is to recommend resolver). For example, given routing sequence $\texttt{A} \rightarrow \texttt{B} \rightarrow \texttt{C}$, there are two edges in  $H_{Trans}$, $R = \{(A,B,1), (B,C,1)\}$ with equal weight. In $H_{Res}$, the edges and weights become $R = \{(A, C, 0.5), (B, C,1)\}$.  The last graph we build is $H_{root}$, which is $H_{Res}$ at root level, \ie aggregating leaf nodes to their roots.

Table~\ref{tab:network} summarizes key statistics of the three networks. The values show that a significant number of routings are originated from a group in the same root. With the same number of nodes, $H_{Trans}$ is more sparse than $H_{Res}$, having only half of the number edges $H_{Res}$ has and smaller average degree. As $H_{Res}$ is more connected, routing models based on it have more options from a node to potential resolvers. The differences in network diameter and average path length measures the longest shortest path and the average path. Both measure are reduced together with the number of nodes at the root level network, indicating a decreased level of interaction at the root level. Average clustering coefficient measures how likely the nodes in a graph tend to form a cluster together. When the coefficient is low ($H_{Trans}$ and $H_{Res}$), the nodes are more likely to form many clusters, based on their connections. The higher score for $H_{root}$ suggests that the nodes are more closely connected.

\section{The UFTR Framework} \label{framework}
\begin{figure*}[t]
	\centering
	\includegraphics[width=1\linewidth]{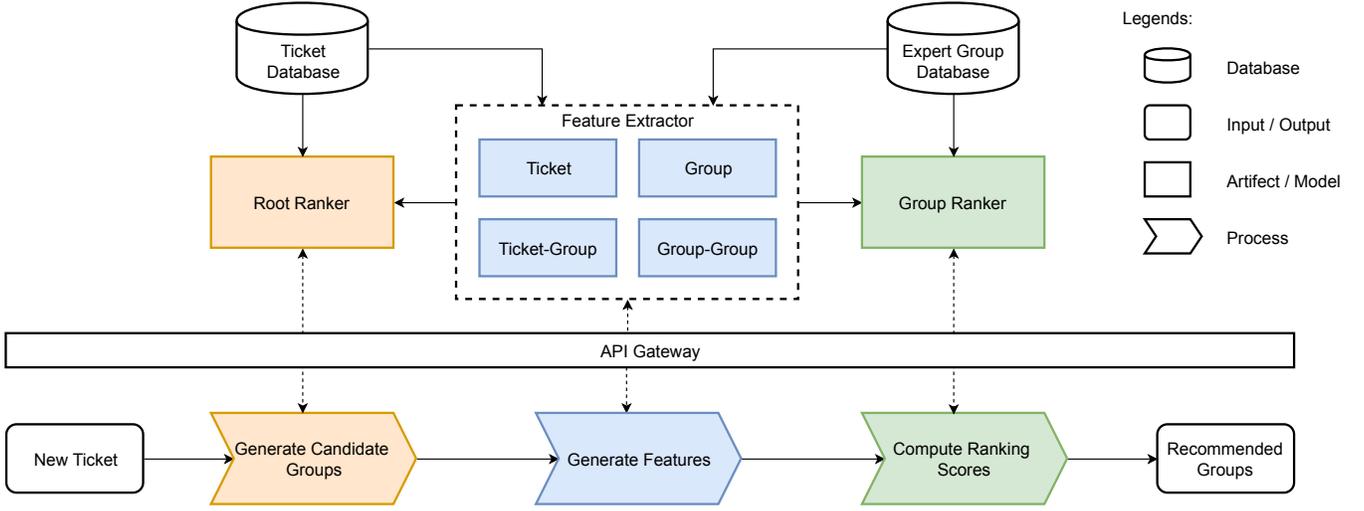}
	\caption{System architecture of the UFTR framework. Components at the upper region are artifacts or models produced from offline training processes, where are not shown in the diagram for simplicity. The lower region contains the main steps in online operation. Online and offline components exchange data through an API gateway.}
	\label{fig:architecture}
\end{figure*}

After analyzing the ticket contents and group networks, we elaborate the UFTR framework and features in greater detail. With UFTR, we attempt to (1) assign an initial group to a ticket, and (2) transfer the ticket to the next group, when the current one failed to resolve it. Technically, given a ticket and its current routing sequence, and  expert group, we want to compute a ranking score indicating the likelihood of the group being the resolver. With the matching scores, we can rank the candidate groups and recommend the top-$n$ to user. The overall architecture of the proposed framework is shown in Figure~\ref{fig:architecture}. The offline training phase consists of three main components: training a \textbf{root ranker} for candidate selection (Section~\ref{candidate selection}). Secondly, given a $\langle$\textit{ticket}, \textit{current routing sequence}, \textit{candidate group}$\rangle$ triplet, multiple artifects and models are produced after trainings in \textbf{feature extractor} (Section~\ref{feature extraction}). Lastly, the \textbf{group ranker} is trained using the features and a binary label to indicate if the candidate group is the resolver of the ticket (Section~\ref{ssec:ranking}). Subsequently, during online processing phase, the trained components are accessed via an API gateway for the routing process, including to candidate groups generation, feature generation, and computing ranking scores.   

\subsection{Root  Ranker} \label{candidate selection}

\begin{figure}[t]%
	\centering
	\includegraphics[width=0.7\linewidth]{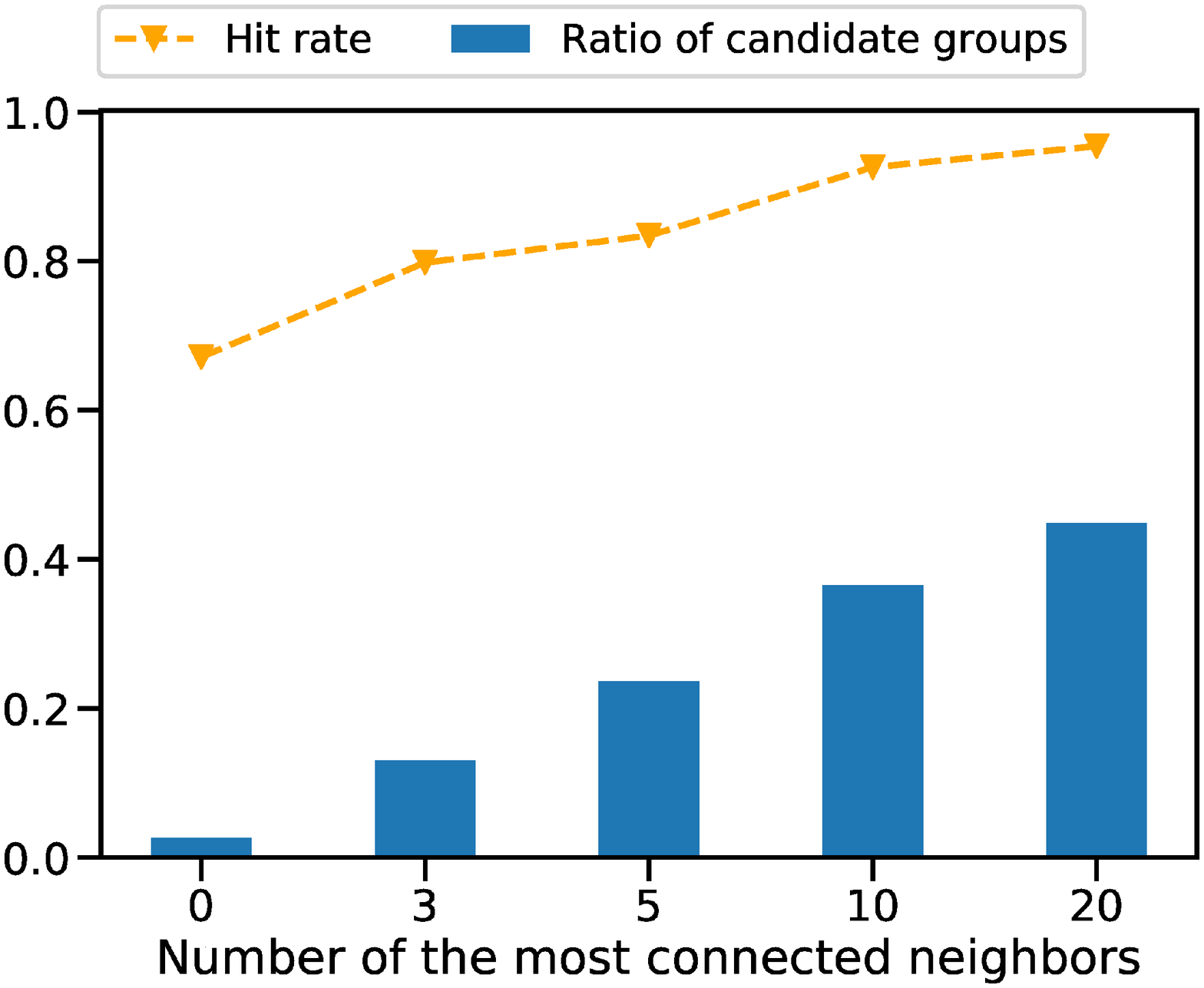}
	\caption{Root level prediction using a unsupervised retrieval-based classifier, expanded with $n$ most connected neighbours in $H_{root}$. The dashed line indicates the number of cases having the resolver among the candidate, over all tickets (\ie hit rate). The blue bar represents the ratio of candidates after expansion, over all expert groups in the network $H_{res}$.}%
	\label{fig:root classification}%
\end{figure}

Given a ticket, a subset of expert groups are first selected as candidate resolvers. As shown in Section~\ref{network analysis}, a majority of the transfers happen between groups within a hierarchy with the same root. 
Intuitively, we attempt to make use of the hierarchy, to filter the candidates for the next stage of processing.

The idea is to select a root and use the groups from it as candidate groups. It can be modeled either as a ticket classification or retrieval task. Training a supervised classifier involves data set preparation, input vectorization, algorithm selection, and model fitting. To focus on the ticket routing problem, we implement a ticket retrieval-based classifier which does not require any training. First, we index every ticket as a document using Apache Lucene~\cite{lucene-web}. Next, using a ticket as query, we retrieve similar tickets using \textsf{MoreLikeThis} similarity query built-in by Lucene. From top-100 returned ticket, we aggregate the corresponding resolver roots by majority voting to select the candidate root.

To increate recall for the ground truth resolver, we expand the candidate set by leveraging the network information in $H_{root}$. Specifically, we find the neighboring root nodes of a predicted root in $H_{root}$. Recall that, the edge weights model the probability of ticket relevance from a root to a resolver root. Sorting the neighboring nodes by weight of edges, we take the top-$n$ roots for candidate expansion. Figure~\ref{fig:root classification} shows the performance of the combined strategy on our data. The figure shows that hit rate is higher than 0.9, with less than 40\% of all groups are covered in candidate sets, when 10 neighboring roots are selected.

\subsection{Feature Extractor} \label{feature extraction}

\begin{table*}[t]
	\small
	\centering
	\caption{Representative features proposed for ticket routing. $\tau$ is the ticket, $E_\tau$ is the entity set contained in $\tau$. $g$ is the candidate group, while $S_\tau$ is the current routing sequence. When a new ticket is being processed, $S_\tau = \emptyset$.}
	\scalebox{0.9}{
		\begin{tabular}{c l l}
			\toprule
			\textbf{Type} & \textbf{Feature} & \textbf{Description} \\
			\midrule
			\multirow{4}{2cm}{\centering Ticket (T)} & $|\tau|$, Clarity($\tau$) & Length of ticket $\tau$ and technical clarity of ticket $\tau$ \\
			& $\sum_{e_i \in E_{\tau}}c(e_i, \tau)$ & Total occurrences of technical entities in ticket $\tau$ \\
			& $\sum_{e_i \in E_{\tau}}c(e_i, \tau) \times 1/|\tau|$ & Ratio of technical entities in ticket $\tau$ \\
			& $\sum_{e_i \in E_{\tau}} IEF(e_i)$ & Sum IEF of technical entities in ticket $\tau$ \\
			
			\midrule
			\multirow{6}{2cm}{\centering Group (G)} & $P_{participant}(g)$, $P_{initial}(g)$, $P_{resolver}(g)$ & Prior probability of $g$  in any routing, as initial group, and as resolver. \\
			& $deg^{-}(g)$, $deg^{+}(g)$, $deg(g)$ & Unweighted in\-degree, out\-degree and total degree of group $g$  \\
			& $deg^{-}_{weighted}(g)$, $deg^{+}_{weighted}(g)$, $deg_{weighted}(g)$ & Weighted in\-degree, out\-degree and total degree of group $g$  \\
			& Harmonic, Closeness, Betweenness centralities & Measures of how close is $g$ to all other groups. \\
			& Eigenvector centrality, PageRank, Hub, Authority & Measures of the quality of $g$ voted by other groups. \\
			& Cluster Coefficient & Cluster coefficient of $g$\\

			\midrule
			\multirow{5}{2cm}{\centering Ticket-Group (TG)} & $P(g|\tau)$ & Likelihood of $g$ being the resolver of $\tau$ \\
			& $\text{cos}_{ent}(\tau,T_g)$, $\text{cos}_{emb}(\tau,T_g)$ & Cosine similarity between ticket $\tau$ and group $g$ and tickets sovled by $g$.   \\
			& Distance($\tau_{emb}$, $g_{emb}$) & Euclidean distance between $\tau$ and $g$ in embedding space.\\
			& QLM, BM25, SDM & Query likelihood model, BM25, and sequential dependency model.  \\
			
			\midrule
			\multirow{2}{2cm}{\centering Group-Group (GG)} & $|S_\tau|$ & length of current routing sequence, 0 for dispatch \\
			& $P_{FMS}(S_\tau \rightarrow g)$, $P_{VMS}(S_\tau \rightarrow g)$, $P_{coll}(S_\tau, g)$ & Transition probabilities and collocations.  \\
			\bottomrule
	\end{tabular}}
	\label{tab:features}
\end{table*}

Language used in tickets are highly specific and technical. Recall that we harness an entity-oriented~\cite{KrisztianBalog2018Entity-orientedSearch} representation approach to bridge the unstructured text to domain vocabularies. We propose four types of features specific for our domain problem, and all features are numerical and normalized to [0,1].

\paratitle{Ticket features}. Ticket features are used to describe a single ticket, independent of any other ticket or group. Only using text content in a ticket, we consider the following ticket statistics as features, \eg length, number of technical entities, and their ratio respective to the whole ticket. From all archived tickets, we compute inverse document frequencies for entities (IEF) and an overall entity language model. Additional features can be derived by summing up the IEF for all entities in a ticket, and technical clarity. 

\paratitle{Group features}. From network of groups  $H_{res}$, we obtain a rich set of network features to characterize each group, \eg the in-, out-, and total degree of a node. These feature are analogous to the transition probabilities in~\cite{Shao2008EfficientMining}. However, we focus on the connection to resolver instead of direct routing. From the network, we can also compute hub and authority scores for the nodes~\cite{kleinberg1999hubs}. Other features include eccentricity, closeness centrality, harmonic centrality, betweenness centrality,  Eigenvector centrality and PageRank~\cite{page1999pagerank}, and cluster coefficient. Besides the network features, we also include the prior probabilities, $log(P_{participant}(g))$, $log(P_{initial}(g))$, and $log(P_{resolver}(g))$.

\paratitle{Ticket-Group features}. This type of feature considers ticket $\tau$ and  group $g$ as a potential resolver. One example is conditional probability $P(g_i|\tau)$. Using the entity-oriented representation, it can be evaluated as \ie $P(g_i|\tau) = p(g)\prod_{e_i \in E_\tau} p(e_i|g)$.  Another feature considers the similarity between $\tau$ and the tickets previously closed by $g$, denoted by $T_g$. For ticket representation, we use technical entities and word embeddings. The former represents $\tau$ and $T_g$ as sparse vectors where each dimension corresponds to a technical entity. The latter is learned by FastText~\cite{bojanowski2016enriching} with skip-gram on our domain corpus. Specifically, we use the average word embedding to represent a ticket and $T_g$. Cosine similarity is computed for both sparse vector and embedding vector representations. Lastly, three IR-based relevance scores are computed by  query likelihood model, BM25, and sequential dependency model, considering $\tau$ as a query and $T_g$ as documents.

\paratitle{Group-Group features}. This type of features are used primarily in transition-based routing models~\cite{Shao2008EfficientMining,Miao2010GenerativeNetworks}. Given a current processing group $g_i$, we estimated the probability of a ticket being transferred to another group $g_j$, from historical data. The \textit{first-order multiple active state search } (FMS) and \textit{variable-order multiple active state search} (VMS) routing models are adopted~\cite{Shao2008EfficientMining}. In addition, we also consider collocation information for two groups, indicating the association of $g_i \in S_\tau$ to $g_j$ if $g_j$ is the resolver. $S_\tau$ is the current routing sequence and $S_\tau$ is empty for a new ticket. In this case, the only information for this group is the probability from dispatch in history; it can also be seen as the prior probability for each group to be the resolver, \ie $P(\emptyset \rightarrow g) = P(g)$.

\subsection{Group Ranker}
\label{ssec:ranking}

Ticket routing problem is similar to a learning-to-rank problem, which is the state-of-the-art model in many retrieval tasks~\cite{Liu2011LearningRetrieval}. Given a query, the objective is to sort a set of documents according to their relevance to the original query. In general,there are three approaches to learning-to-rank, \ie pointwise, pairwise, and listwise~\cite{KrisztianBalog2018Entity-orientedSearch}. In pointwise approach, the problem reduces to a regression task where the objective is to learn a scoring function for each query-document pair, independent of other documents in the list. The pairwise approach takes pairs of document, learning a model to classify if one should be ranked ahead of the other. Listwise approach considers the whole list of documents as input, attempting to optimize the ordering directly. 

\begin{algorithm}

	\SetKwInOut{Input}{input}
	\SetKwInOut{Output}{output}
	
	\Input{Ticket $\tau$, \newline expert group network $H_{root}, H_{trans}$ }
	\Output{Candidates groups $C_{group}$}

	$C_{group} \leftarrow \emptyset$\; 
	
	$\hat{r_\tau} \leftarrow$ Classify $\tau$ using retrieval-based classifier\;

	$R_\tau \leftarrow \{r : r \in H_{root}, \hat{r_\tau}r \in H_{root}.edges \}$\;
	
	$C_{root} \leftarrow$ top-$N$ ranked $r$ in $R_\tau$ by $weight(\hat{r_\tau}r)$\;
	
	\ForEach(\tcp*[f]{iterate roots}){$r$ in $C_{root} $ }
	{
		$r' \leftarrow$ Locate corresponding $r$ in $H_{trans}$\;
		
		$G_c \leftarrow \{r : r \in H_{trans}, r.isDescendentOf(r')\}$\;
		
		$C_{group} \leftarrow C_{group} \bigcup G_c$\;
	}

	\KwRet {$C_{group}$}

	\caption{Candidate Groups Generation}
	\label{algo:candidate} 
\end{algorithm}

\begin{algorithm}[t]
	\SetKwInOut{Input}{input}
	\SetKwInOut{Output}{output}
	
	\Input{Ticket $\tau$, \newline current routing sequence $S_t$, \newline expert group network $H_{root}, H_{trans}$ \newline Candidates groups $C$}
	\Output{Groups $R_g$}
	
	$scores \leftarrow \emptyset$\;
	 
	\ForEach(\tcp*[f]{iterate candidates}){$g$ in $C_{group} $ }
	{
		Compute feature vector for $(\tau, S_t, g)$\;

		$scores.append((g, score(\tau, S_t, g)))$\;
	}
	
	$R_g \leftarrow$ top-$k$ ranked $g$ in $scores$ by $score(\tau, S_t, g)$\;
	
	\KwRet {$R_g$}

	\caption{Resolver Candidate Ranking}
	\label{algo:ranking} 
\end{algorithm}

In ticket routing, query is a ticket with routing sequence, while the document is expert group. To create training instances, for a $\langle$ ticket, current routing sequence, candidate group $\rangle$ triplet, the features are generated from the feature extractor module. Note the candidate group is from the ground truth routing sequence (positive samples) or random sampling (negative samples). From each triplet we create a training instance {($\mathbf{x}_{\tau, S_t, g_{t+1}}, r_{\tau, S_t, g_{t+1}}$)}. In which $\mathbf{x}_{\tau, S_t, g_{t+1}}$ is a multi-dimensional feature vector, and $r_{\tau, S_t, g_{t+1}} \in \mathbb{R}: 0 \leq r_{\tau, S_t, g_{t+1}} \leq 1$ is the relevance score.  Note that, a ticket with multiple routing steps are converted to multiple training instances, the non-resolver groups have discounted relevance score as in the way we construct  $H_{Res}$. Specifically, we assign the distance-based scores and generate positive training samples by pairing a ticket with every group in its routing sequence. Each pair is assigned with a value $(0, 1]$ relative to its distance to the resolver; 1 is assigned to resolver and the score reduces by half as one step further from the resolver. For every positive sample, we randomly select an expert group that was not involved in the routing as a negative sample, \ie setting the relevant score to $0$.

In this paper, we implemented both pointwise and pairwise learning-to-rank for group ranker. Specifically, we train a Random Forest Regressor (RFR)~\cite{Breiman:2001:RF:570181.570182} with 200 regression trees and bootstrapping for the former approach. The benefit of such model is its explainability. The model scores each features by their impact to the final scoring. For pairwise approach, we employ LambdaMART~\cite{wu2010adapting}, one of the state-of-the-art algorithms for learning-to-rank. It also leverages regression trees, with gradient boosting. The gradients are approximated from the evaluation metric in ranking, \eg normalized discounted cumulative gain (NDCG). As the listwise approach is computationally challenging to implement, we leave it for future works.

\section{Experiment}
We evaluate the performance of UFTR in ticket routing task. In addition, we also study its performance in resolver ranking, and investigate contribution of individual features to the results.

\subsection{Experiment Setup} \label{experiement}

\paratitle{Dataset.}  We select archived tickets with routing sequences for evaluation. From all tickets used for analysis in Section~\ref{data analysis}, we randomly select 58,670 (11.7\%) tickets for training.   From the selected training tickets, we generate 150,000 training instances with positive and negative ratio of 1:1, as the training set. The remaining tickets are grouped by routing steps and 500 are randomly selected from each group having $n$ ($n\leq 4$) steps. Effectively, we created four test sets, S1, S2, S3, and S4.

\paratitle{Evaluation metrics.} For each test ticket, we simulate a routing process assisted by our ranking model. Initially, the routing sequence is empty and the model recommends a list of candidate groups as potential resolvers. Depends on the evaluation metric to be detailed, if the ticket is considered not resolved, routing continues. The performance is evaluation using the following metrics:  Mean Steps to Resolver (\textbf{MSTR}), Resolution Rate (\textbf{RR}), and  Mean Average Distance to Resolver (\textbf{MADR}@$k$).

Mean Steps to Resolver (MSTR) is the average length of test routing sequences. The closer the value is to 1, the more effective a system is. In experiment, for each test ticket, we consider the top-1 ranked group recommended by a system in each routing step. The recommendation is successful only if the ground truth resolver is correctly recommended. If one step fails, the input to the next step recommendation is the ground truth routing sequence or the previous recommended routing sequence. For each test set, an effective routing solution should be lower than the average archived routing steps. We assume no ticket would take more than 10 steps in this experiment. Resolution Rate (RR), measures the ratio of resolved test tickets within $n$ recommended steps. Given $n$, more tickets is resolved, the better a system is. Mean Average Distance to Resolver (MADR@$k$)~\cite{Han2017MeanNetwork} scores each routing step with respect to the ground truth routing sequence. Formally,

\begin{equation}
MADR(\mathcal{T}) = \frac{\sum_{t\in \mathcal{T}}\phi(\tau)}{|\mathcal{T}|}
\end{equation}
where,
\begin{equation}
\phi(t)=\frac{\sum_{g \in \hat S_\tau}\varphi(g)}{|\hat S_\tau|}
\label{eqn:ags}
\end{equation}

where $\varphi(g)$ is a distance-based group scoring function $\varphi(g)=\frac{1}{2^{\triangle (g, g_{resolver})}}$, $\triangle$ represents ``distance between".
At each routing step, a system generates top-$k$ recommendations. The routing is successful if the resolver is among the top-$k$. Otherwise, the next group in the ground truth routing sequence is added to the routing sequence. In this way, the process incorporates more human decisions in the archived tickets.

\paratitle{Baseline models}.  As discussed in Section~\ref{related works}, only Two-stage Expert Recommendation (TER) framework~\cite{Xu2018ExpertRouting} that uses four types of features is comparable to UFTR. We re-implement TER as a baseline. The first stage selects the group with minimal distance to the ticket in embedding space. The group representations are learned using the model proposed by~\cite{Han2016DistributedExpertise}, using all the tickets a group has resolved as positive samples and randomly selected negative samples. The second step follows the transition based model as proposed by~\cite{Shao2008EfficientMining}, using group-group transition probabilities modeled in $H_{Trans}$. By fixing the first step  as proposed and only changing the second stage, we have the following three routing baselines\footnote{We also experimented with the generative model proposed in~\cite{Miao2010GenerativeNetworks}. Unfortunately, we are unable to reproduce sensible results as the model is untractable on our data. The number of parameters to be optimized are $\mathcal{O}(VTE^2G^2)$, where $V$ is the vocabulary size, $T$ is size of training set, $E$ is the number of edges in the network, and $G$ is the group count.}.

\textbf{TER-FM} is a first-order memoryless search model using a greedy approach. The next group $g_{t+1}$ is selected using
$$g_{t+1} = \text{argmax}_g P(g|g_{t}), \forall g \in \mathcal{G},$$ where $g_{t}$ is the current group.

\textbf{TER-FMS} is a first-order multiple active state search model, considering groups in the current routing sequence instead of only the current group. The next group $g_{t+1}$ is selected using
$$g_{t+1} = \text{argmax}_g P(g|g_r), \forall g_r \in S_{\tau}, g \in \mathcal{G} \backslash S_{\tau},$$ where  $S_{\tau}$ is the set of groups that is in the current routing sequence.

\textbf{TER-VMS} is a variable-order search algorithm, using a higher order Markov model. The next group $g_{t+1}$ is selected using
$$g_{t+1} = \text{argmax}_g P(g|S_{k}), \forall g \in \mathcal{G} \backslash S_{\tau}, S_k \subseteq S_{\tau},$$ where $S_k$ is a subset of the current routing sequence.

\subsection{Evaluation on Ticket Routing }

\begin{figure*}[t]
	\centering
	\includegraphics[width=1\textwidth]{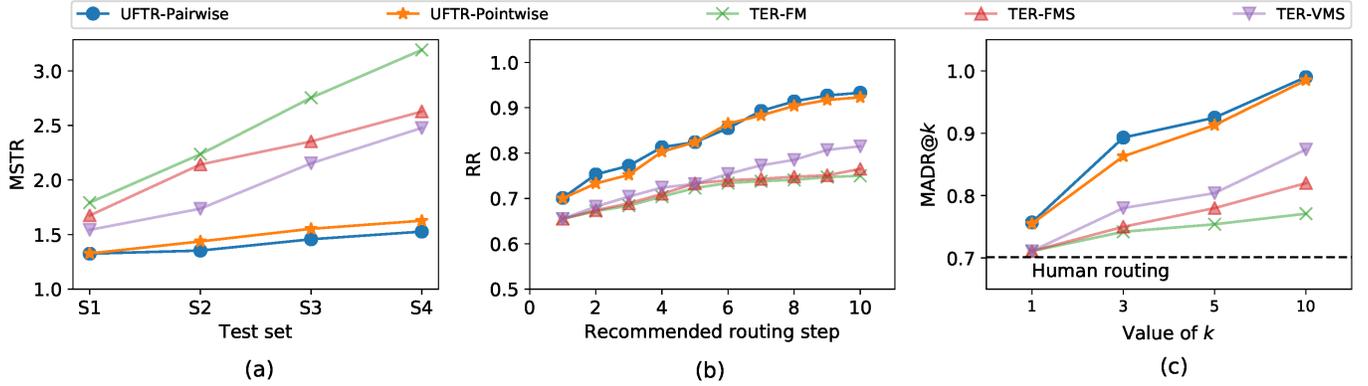}
	\caption{Performance comparisons of UFTR with baselines in (a) MSTR, (b) RR, and (c) MADR@$k$}.
	\label{fig:performance}
\end{figure*}

Ideally,  UFTR can be used to rank all candidate groups with the trained model on every route step. For the purpose of experimentation, we adopt the \textit{leave-one-out} test scheme that is widely used in evaluating recommendation systems~\cite{elkahky2015multi,he2016fast}. Specifically, for each test ticket, other than the resolver group, we select another 49 other groups to form 50 pairs for feature generation, and ranking. These 49 groups include groups from the archived routing sequence, and rest are randomly sampled from the candidate set produced by root prediction and expansion with 10 neighbor roots (see Section~\ref{candidate selection}).

Figure~\ref{fig:performance} reports the results of UFTR against baselines, in three metrics. In Figure~\ref{fig:performance} (a), UFTR-Pairwise is the best performing model across all four test sets in terms of MSTR, followed by UFTR-Pointwise. For S1, the scores are dominated by the results of initial group assignments. Our root ranker with graph-based expansion performs better than the TER-based systems which use a distance-based initial group assignment approach. In S2, UFTR-Pairwise and UFTR-Pointwise reduces MSTR from 2 to 1.4 and 1.3 (the lower the better), respectively. TER-VMS also cuts on the archived routing length by 0.4,  being the best performing TER-based model. MSTRs yield from the other two baselines are even longer than those of the archived routing, \ie human routing. In S3 and S4, all systems exhibit some degree of MSTR reduction, with UFTR-Pairwise performed the best. Among the TER-based systems, the TER-VMS model performs slightly better than TER-FM and TER-FMS.

Figure~\ref{fig:performance}(b) shows the changes in resolution rate as routing steps increase from 1 to 10. A monotonic increase of RR is observed from all systems. The increase of RR for UFTR-based methods are more  steep, compared to the TER-based methods. However, the performance of pairwise and pointwise UFTR methods are comparable, with the former slightly higher at lower routing steps.

Figure~\ref{fig:performance}(c) reports MADR. A human routing baseline is computed by averaging the scores of ground truth routing sequences. The $x$-axis is the value of $k$, which is the top positions to consider a hit if the ground truth resolver is in. As $k$ increases, the criteria of finding a resolver is relaxed. UFTR-Pairwise has the steepest incline as $k$ increases from 1 to 10, approaching the theoretical best score of 1. UFTR-Point is not too far behind. The baselines show slower improvement over human routing, among which TER-VMS is better than the others. After all, UFTR-based methods are more effective than all baselines for all metrics, while pairwise approach is slightly better than pointwise.

\subsection{Evaluation on Resolver Ranking }

\begin{table*}[t]
	\centering
	\caption{ Hit Rate (HR) of UFTR-Pariwise for the task of resolver ranking on four test sets.}
	\label{tab:hit rate}
	\begin{tabular}{l | ccc |ccc |ccc | ccc}
		\toprule
		& \multicolumn{3}{c|}{\textbf{Test set S1}}& \multicolumn{3}{c|}{\textbf{Test set  S2}} & \multicolumn{3}{c|}{\textbf{Test set  S3}} & \multicolumn{3}{c}{\textbf{Test set  S4}}  \\
		\textbf{Features} & \textbf{HR@1} & \textbf{HR@3} & \textbf{HR@5} & \textbf{HR@1} & \textbf{HR@3} & \textbf{HR@5} & \textbf{HR@1} & \textbf{HR@3} & \textbf{HR@5} & \textbf{HR@1} & \textbf{HR@3} & \textbf{HR@5} \\
		\midrule
		All Features & 0.792 & 0.950 & 0.972   & 0.674 & 0.942 & 0.978   & 0.672 & 0.938 & 0.978   & 0.700 & 0.936 & 0.970 \\
		- less T & 0.720 &	0.938 & 0.968 & 0.602 &	0.928 &	0.976 &	0.566 &	0.884 &	0.960 &	0.664 &	0.906 &	0.954 \\
		- less G & 0.740 & 0.946 & 0.948 &	0.592 &	0.928 &	0.974 &	0.580 & 0.898 & 0.960 & 0.692 & 0.918 & 0.968 \\
		- less TG & 0.290 & 0.494	& 0.588 & 0.088 & 0.226 & 0.362 & 0.106 & 0.234 & 0.372 & 0.096	& 0.218 &	0.368\\
		- less GG & 0.746	& 0.948	& 0.972	& 0.620	& 0.932	& 0.974	& 0.650	& 0.920	& 0.960	& 0.708	& 0.916	& 0.970 \\
		\bottomrule
	\end{tabular}
\end{table*}

\begin{table}[t]
	\centering
	\caption{Most important features and their types, ranked by importance scores from the Random Forest Regressor.}
	\begin{tabular}{l l c c}
		\toprule
		\textbf{Rank} & \textbf{Feature} & \textbf{Importance} & \textbf{Type}  \\
		\midrule
		1 & QLM & 0.351 & TG \\
		2 & $P(g|\tau)$ & 0.273 & TG \\
		3 & BM25 & 0.013 & TG \\
		4 & Clarity & 0.011 & T \\
		5 & $\sum_{e_i \in E_{\tau}}c(e_i, \tau) \times 1/|\tau|$ & 0.009 & T \\
		6 & $cos_{ent}$ &  0.008 & TG \\
		7 & $cos_{emb}$, $\sum_{e_i \in E_{\tau}}c(e_i, \tau)$ &  0.007 & TG, T \\
		8 & $P_{resolver}(g)$ & 0.006 & G \\
		9 & ClusCoef, $P_{VMS}(S_\tau \rightarrow g)$ & 0.005 & G, GG \\
		10 & $|\tau|$, $deg^+_{weighted}$ & 0.004 & T, G \\
		\bottomrule
	\end{tabular}
	\label{tab: feature importance}
\end{table}

In this experiment, we evaluate the performance of our best model and the importance of individual features.
For each test set, Table~\ref{tab:hit rate} presents the hit rates of resolver within the top-$k$ ranking position, $k \in {1, 3, 5}$. The first model uses all features, and the rest of four models are trained by removing one type of features each time. The hit rates increase significantly as $k$ increases, reaching over 0.97 when all features are used when $k=5$. It shows that our model is capable of ranking the resolver to the top positions among non-resolvers. The hit rates on S2 is slightly better than those on S3, but came short comparing to the results on S1, especially at top-1 position. As S2 and S3 contain tickets that are resolved by 2 and 3 steps, respectively, the non-resolver groups may be similar to the resolver group in many aspects. In this case, the model is unable to identify the resolver accurately. However, on S4, the top-1 hit rate is higher than both S2 and S3, only after S1. This suggests that the longer routing sequences in the archived tickets may be less necessary.

For models trained using less feature type, the hit rate decreases for all $k$ values except for HR@1 for S4. This could be due to the noise brought by network features. The most decline is observed in the rest of HR@1 measures. Noticeably, when  Ticket-Group (TG) features are excluded, the hit rates drop drastically. The results of this ablation study show that Ticket-Group features are the most important among all feature types. Previous study~\cite{Xu2018ExpertRouting} concluded that ticket content features are more important than routing sequence in routing task. Though the discrepancy could be due to differences in data distribution and problem setup, it is worth to investigate more features to model the interaction of a ticket and an expert group in future work.

\subsection{Feature Importance}

Knowing that  Ticket-Group features are the most important type of features, we want to further understand individual feature's contribution to the final results. The RFR model is able to output a weight indicating feature importance.  Table~\ref{tab: feature importance} lists the top-10 features that are ranked by this weight, among all features across the four types. The three most weighted features are all from the Ticket-Group type. The score by query likelihood model, the conditional probability of the group given a ticket, and score by BM25 are with great importance. Among features from other types, ticket clarity appears to be the most useful. On the other hand, the Group features carry less weights in general. In Group-Group features, the probability of being a resolver, and the transition probability under the VMS strategy, are relatively more important.

\section{Conclusion and Future Work} \label{conclusion}
In this work, we proposed the UFTR, a unified framework for ticket routing. We first analyze the contents and group characteristics of half a million archived tickets. Then we propose four types of features to represent different aspects of the routing problem. The features are used to train group ranking models. Through experiments, we show that the UFTR with pairwise ranking model beats other approaches in three routing evaluation metrics. It outperforms its pointwise variation, as well as baselines using two-staged modeling. In an ablation study, we observe that the Ticket-Group features, which represent the interaction between a ticket with a group, are the most important features. Unlike previous solutions that use different models for initial group assignment and transfer, the UFTR uses a single model for both subproblems. In future research, we plan to further investigate the Ticket-Group interactions, as well as the more sophisticated ranking methods (\eg listwise) to solve the ticket routing problem.

\bibliographystyle{IEEEtran}
\bibliography{ticket-routing}

\begin{thebibliography}{10}
\providecommand{\url}[1]{#1}
\csname url@samestyle\endcsname
\providecommand{\newblock}{\relax}
\providecommand{\bibinfo}[2]{#2}
\providecommand{\BIBentrySTDinterwordspacing}{\spaceskip=0pt\relax}
\providecommand{\BIBentryALTinterwordstretchfactor}{4}
\providecommand{\BIBentryALTinterwordspacing}{\spaceskip=\fontdimen2\font plus
\BIBentryALTinterwordstretchfactor\fontdimen3\font minus
  \fontdimen4\font\relax}
\providecommand{\BIBforeignlanguage}[2]{{%
\expandafter\ifx\csname l@#1\endcsname\relax
\typeout{** WARNING: IEEEtran.bst: No hyphenation pattern has been}%
\typeout{** loaded for the language `#1'. Using the pattern for}%
\typeout{** the default language instead.}%
\else
\language=\csname l@#1\endcsname
\fi
#2}}
\providecommand{\BIBdecl}{\relax}
\BIBdecl

\bibitem{galup2009overview}
S.~D. Galup, R.~Dattero, J.~J. Quan, and S.~Conger, ``An overview of it service
  management,'' \emph{Communications of the ACM}, vol.~52, no.~5, pp. 124--127,
  2009.

\bibitem{Shao2008EasyTicket:Resolution}
Q.~Shao, Y.~Chen, S.~Tao, X.~Yan, and N.~Anerousis, ``Easyticket: a ticket
  routing recommendation engine for enterprise problem resolution,''
  \emph{VLDB}, vol.~1, no.~2, pp. 1436--1439, 2008.

\bibitem{Miao2010GenerativeNetworks}
G.~Miao, L.~E. Moser, X.~Yan, S.~Tao, Y.~Chen, and N.~Anerousis, ``Generative
  models for ticket resolution in expert networks,'' in \emph{KDD}, 2010, pp.
  733--742.

\bibitem{Moharreri2016MotivatingManagement}
K.~Moharreri, J.~Ramanathan, and R.~Ramnath, ``Motivating dynamic features for
  resolution time estimation within it operations management,'' in \emph{IEEE
  Big Data}, 2016, pp. 2103--2108.

\bibitem{Xu2018ExpertRouting}
J.~Xu and R.~He, ``Expert recommendation for trouble ticket routing,''
  \emph{Data \& Knowledge Engineering}, vol. 116, pp. 205--218, 2018.

\bibitem{Motahari-Nezhad2011}
H.~R. Motahari-Nezhad and C.~Bartolini, ``Next best step and expert
  recommendation for collaborative processes in it service management,'' in
  \emph{Business Process Management}, 2011, pp. 50--61.

\bibitem{BShivaliAgarwalRenukaSindhgatta2012}
S.~Agarwal, R.~Sindhgatta, and B.~Sengupta, ``Smartdispatch: enabling efficient
  ticket dispatch in an it service environment,'' in \emph{KDD}, 2012, pp.
  1393--1401.

\bibitem{Zeng2017KnowledgeData}
C.~Zeng, W.~Zhou, T.~Li, L.~Shwartz, and G.~Y. Grabarnik, ``Knowledge guided
  hierarchical multi-label classification over ticket data,'' \emph{IEEE Trans.
  on Network and Service Management}, vol.~14, no.~2, pp. 246--260, 2017.

\bibitem{Han2018VerticalNetworks}
J.~Han and M.~Akbari, ``{Vertical Domain Text Classification: Towards
  Understanding IT Tickets Using Deep Neural Networks},'' in \emph{AAAI}, 2018,
  pp. 8202--8203.

\bibitem{Shao2008EfficientMining}
Q.~Shao, Y.~Chen, S.~Tao, X.~Yan, and N.~Anerousis, ``Efficient ticket routing
  by resolution sequence mining,'' in \emph{KDD}, 2008, pp. 605--613.

\bibitem{Khan2009AIM-HI:Delivery}
A.~Khan, H.~Jamjoom, and J.~Sun, ``Aim-hi: a framework for request routing in
  large-scale it global service delivery,'' \emph{IBM Journal of Research and
  Development}, vol.~53, no.~6, p.~4, 2009.

\bibitem{Sun2010}
P.~Sun, S.~Tao, X.~Yan, N.~Anerousis, and Y.~Chen, ``Content-aware resolution
  sequence mining for ticket routing,'' in \emph{Business Process Management},
  2010, pp. 243--259.

\bibitem{Chen2010AssessingNetworks}
Y.~Chen, S.~Tao, X.~Yan, N.~Anerousis, and Q.~Shao, ``Assessing expertise
  awareness in resolution networks,'' in \emph{ASONAM}, 2010, pp. 128--135.

\bibitem{Miao2012UnderstandingNetworks}
G.~Miao, S.~Tao, W.~Cheng, R.~Moulic, L.~E. Moser, D.~Lo, and X.~Yan,
  ``Understanding task-driven information flow in collaborative networks,'' in
  \emph{WWW}, 2012, pp. 849--858.

\bibitem{Sun2014AnalyzingNetworks}
H.~Sun, M.~Srivatsa, S.~Tan, Y.~Li, L.~M. Kaplan, S.~Tao, and X.~Yan,
  ``Analyzing expert behaviors in collaborative networks,'' in \emph{KDD},
  2014, pp. 1486--1495.

\bibitem{Han2016DistributedExpertise}
F.~Han, S.~Tan, H.~Sun, M.~Srivatsa, D.~Cai, and X.~Yan, ``Distributed
  representations of expertise,'' in \emph{SIAM Data Mining}, 2016, pp.
  531--539.

\bibitem{Moharreri2015}
K.~Moharreri, J.~Ramanathan, and R.~Ramnath, ``Recommendations for achieving
  service levels within large-scale resolution service networks,'' in \emph{ACM
  India Conference}, 2015, pp. 37--46.

\bibitem{Moharreri2016ProbabilisticNetworks}
------, ``Probabilistic sequence modeling for trustworthy it servicing by
  collective expert networks,'' in \emph{IEEE COMPSAC}, vol.~1, 2016, pp.
  379--389.

\bibitem{Diao2009}
Y.~Diao, H.~Jamjoom, and D.~Loewenstern, ``Rule-based problem classification in
  it service management,'' in \emph{IEEE Cloud Computing}, 2009, pp. 221--228.

\bibitem{Simoudis1992}
E.~Simoudis, ``Using case-based retrieval for customer technical support,''
  \emph{{IEEE} Expert}, vol.~7, no.~5, pp. 7--12, 1992.

\bibitem{Kang2014AKnowledge}
Y.-B. Kang, S.~Krishnaswamy, and A.~Zaslavsky, ``A retrieval strategy for
  case-based reasoning using similarity and association knowledge,'' \emph{IEEE
  transactions on cybernetics}, vol.~44, no.~4, pp. 473--487, 2014.

\bibitem{MotahariNezhad2011AnalyticsFlows}
H.~R.~M. Nezhad, C.~Bartolini, and P.~Joshi, ``Analytics for similarity
  matching of it cases with collaboratively-defined activity flows,'' in
  \emph{ICDE Workshops}, 2011, pp. 273--278.

\bibitem{Han2018TowardsTickets}
J.~Han, K.~H. Goh, A.~Sun, and M.~Akbari, ``Towards effective extraction and
  linking of software mentions from user-generated support tickets,'' in
  \emph{CIKM}, 2018, pp. 2263--2271.

\bibitem{gdpr-web}
\BIBentryALTinterwordspacing
The eu general data protection regulation (gdpr). [Online]. Available:
  \url{https://eugdpr.org/}
\BIBentrySTDinterwordspacing

\bibitem{Cronen-Townsend2002PredictingPerformance}
S.~Cronen-Townsend, Y.~Zhou, and W.~B. Croft, ``Predicting query performance,''
  in \emph{SIGIR}, 2002, pp. 299--306.

\bibitem{lucene-web}
Lucene: Ultra-fast search library and server.

\bibitem{KrisztianBalog2018Entity-orientedSearch}
K.~Balog, \emph{Entity-oriented Search}.\hskip 1em plus 0.5em minus 0.4em\relax
  Springer, 2018.

\bibitem{kleinberg1999hubs}
J.~M. Kleinberg, ``Hubs, authorities, and communities,'' \emph{ACM computing
  surveys}, vol.~31, no. 4es, p.~5, 1999.

\bibitem{page1999pagerank}
L.~Page, S.~Brin, R.~Motwani, and T.~Winograd, ``The pagerank citation ranking:
  Bringing order to the web.'' Stanford InfoLab, Tech. Rep., 1999.

\bibitem{bojanowski2016enriching}
P.~Bojanowski, E.~Grave, A.~Joulin, and T.~Mikolov, ``Enriching word vectors
  with subword information,'' \emph{arXiv:1607.04606}, 2016.

\bibitem{Liu2011LearningRetrieval}
T.-Y. Liu, \emph{{Learning to Rank for Information Retrieval}}.\hskip 1em plus
  0.5em minus 0.4em\relax Springer Science \& Business Media, 2011.

\bibitem{Breiman:2001:RF:570181.570182}
L.~Breiman, ``Random forests,'' \emph{Machine Learning}, vol.~45, no.~1, pp.
  5--32, 2001.

\bibitem{wu2010adapting}
Q.~Wu, C.~J. Burges, K.~M. Svore, and J.~Gao, ``Adapting boosting for
  information retrieval measures,'' \emph{Information Retrieval}, pp. 254--270,
  2010.

\bibitem{Han2017MeanNetwork}
J.~Han and A.~Sun, ``Mean average distance to resolver: An evaluation metric
  for ticket routing in expert network,'' in \emph{IEEE ICSME}, 2017, pp.
  594--602.

\bibitem{elkahky2015multi}
A.~M. Elkahky, Y.~Song, and X.~He, ``A multi-view deep learning approach for
  cross domain user modeling in recommendation systems,'' in \emph{WWW}, 2015,
  pp. 278--288.

\bibitem{he2016fast}
X.~He, H.~Zhang, M.-Y. Kan, and T.-S. Chua, ``Fast matrix factorization for
  online recommendation with implicit feedback,'' in \emph{SIGIR}, 2016, pp.
  549--558.

\end{thebibliography}

\vspace{-1cm}

\end{document}